\documentclass{article}

\PassOptionsToPackage{numbers, compress}{natbib}
\usepackage{iclr2017_conference}

\usepackage[utf8]{inputenc} 
\usepackage[hidelinks]{hyperref}       
\usepackage{url}            
\usepackage{booktabs}       
\usepackage{amsfonts}       
\usepackage{nicefrac}       
\usepackage{microtype}      
\usepackage{graphicx}
\usepackage{amsmath}
\usepackage{wrapfig}
\usepackage{soul}

\newcommand{\figref}[1]{Fig.~\ref{#1}}

\newcommand{\sref}[1]{Sect.~\ref{#1}}

\renewcommand{\vec}{\mathbf}

\title{Video Pixel Networks}

\author{
  {Nal Kalchbrenner},
  {A\"aron van den Oord},
  {Karen Simonyan} \vspace{0.08cm} \\ 
  \hspace{0.02cm} \textbf{Ivo Danihelka},
  \textbf{Oriol Vinyals},
  \textbf{Alex Graves},
  \textbf{Koray Kavukcuoglu} \\ \\ 
 \hspace{0.02cm} \texttt{\{nalk,avdnoord,simonyan,danihelka,vinyals,gravesa,korayk\}@google.com} \vspace{0.08cm}  \\
 \hspace{0.02cm} Google DeepMind, London, UK
}

\begin{document}

\maketitle

\begin{abstract}

We propose a probabilistic video model, the Video Pixel Network (VPN), that estimates the discrete joint distribution of the raw pixel values in a video. The model and the neural architecture reflect  the time, space and color structure of video tensors and encode it as a four-dimensional dependency chain. The VPN approaches the best possible performance on the Moving MNIST benchmark, a leap over the previous state of the art, and the generated videos show only minor deviations from the ground truth. The VPN also produces detailed samples on the action-conditional Robotic Pushing benchmark and generalizes to the motion of novel objects. 
\end{abstract}

\section{Introduction}

Video modelling has remained a challenging problem due to the complexity and ambiguity inherent in video data. Current approaches range from mean squared error models based on deep neural networks~\citep{ICML-2015-SrivastavaMS,DBLP:conf/nips/OhGLLS15}, to models that predict quantized image patches~\citep{DBLP:journals/corr/RanzatoSBMCC14}, incorporate motion priors~\citep{DBLP:journals/corr/PatrauceanHC15,DBLP:journals/corr/FinnGL16} or use adversarial losses~\citep{DBLP:journals/corr/MathieuCL15,scenedynamics}. Despite the wealth of approaches, future frame predictions that are free of systematic artifacts (e.g. blurring) have been out of reach even on relatively simple benchmarks like Moving MNIST~\citep{ICML-2015-SrivastavaMS}.

We propose the Video Pixel Network (VPN), a generative video model based on deep neural networks, that reflects the factorization of the joint distribution of the pixel values in a video. The model encodes the four-dimensional structure of video tensors and captures dependencies in the time dimension of the data, in the two space dimensions of each frame and in the color channels of a pixel. 
This makes it possible to model the stochastic transitions locally from one pixel to the next and more globally from one frame to the next without introducing independence assumptions in the conditional factors. The factorization further ensures that the model stays fully tractable; the likelihood that the model assigns to a video can be computed exactly.  The model operates on pixels without preprocessing and predicts discrete multinomial distributions over raw pixel intensities, allowing the model to estimate distributions of any shape. 

The architecture of the VPN consists of two parts: resolution preserving CNN encoders and PixelCNN decoders \citep{van2016pixel}. The CNN encoders preserve at all layers the spatial resolution of the input frames in order to maximize representational capacity. The outputs of the encoders are combined over time with a convolutional LSTM that also preserves the resolution \citep{hochreiter1997long,DBLP:conf/nips/ShiCWYWW15}. The PixelCNN decoders use masked convolutions to efficiently capture space and color dependencies and use a softmax layer to model the multinomial distributions over raw pixel values. The network uses dilated convolutions in the encoders to achieve larger receptive fields and better capture global motion. The network also utilizes newly defined multiplicative units and corresponding residual blocks.

We evaluate VPNs on two benchmarks. The first is the Moving MNIST dataset \citep{ICML-2015-SrivastavaMS} where, given 10 frames of two moving digits, the task is to predict the following 10 frames. In~\sref{sec:moving_mnist} we show that the VPN achieves 87.6 nats/frame, a score that is near the lower bound on the loss (calculated to be 86.3 nats/frame); this constitutes a significant improvement over the previous best result of 179.8 nats/frame~\citep{DBLP:journals/corr/PatrauceanHC15}.

The second benchmark is the Robotic Pushing dataset \citep{DBLP:journals/corr/FinnGL16} where, given two natural video frames showing a robotic arm pushing objects,

the task is to predict the following 18 frames. We show that 

the VPN not only generalizes to new action sequences with objects seen during training, but also to new action sequences involving \emph{novel} objects not seen during training. Random samples from the VPN preserve remarkable detail throughout the generated sequence. 
We also define a baseline model that lacks the space and color dependencies. This lets us see that the latter dependencies are crucial for avoiding systematic artifacts in generated videos. 

\begin{figure}
\vspace{-0.5cm}
\centering
\includegraphics[width=0.75\linewidth]{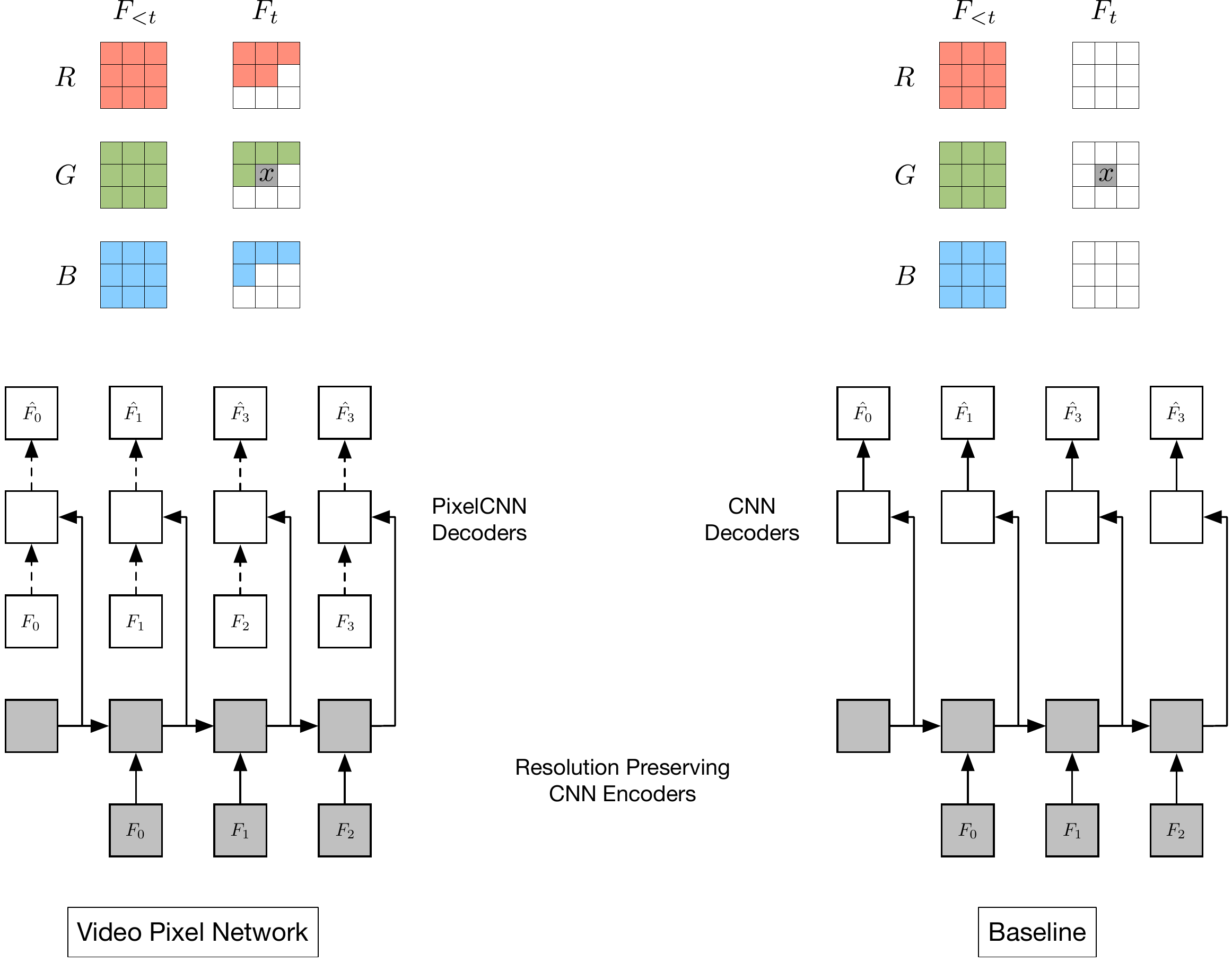}
\caption{
Dependency map (top) and neural network structure (bottom) for the VPN (left) and the baseline model (right).
}
\vspace{-0.5cm}
\label{fig:dependencies}
\end{figure}

\section{Model}
\label{sec:factor}
In this section we define the probabilistic model implemented by Video Pixel Networks.
Let a video $\vec{x}$ be a four-dimensional tensor of pixel values $\vec{x}_{t,i,j,c}$, where the first (temporal) dimension $ t \in \{0,...,T\}$ corresponds to one of the frames in the video, the next two (spatial) dimensions $i,j \in \{0,...,N\}$ index the pixel at row $i$ and column $j$ in frame $t$, and the last dimension $c \in \{R, G, B\}$ denotes one of the three RGB channels of the pixel. 
We let each $\vec{x}_{t,i,j,c}$ be a random variable that takes values from the RGB color intensities of the pixel. 

By applying the chain rule to factorize the video likelihood $p(\vec{x})$ as a product of conditional probabilities, we can model it in a tractable manner and without introducing independence assumptions:

\begin{equation}
p(\vec{x}) = \prod_{t=0}^{T}\prod_{i=0}^N\prod_{j=0}^Np(\vec{x}_{t,i,j,B} | \vec{x}_{<}, \vec{x}_{t,i,j,R}, \vec{x}_{t,i,j,G})\ p(\vec{x}_{t,i,j,G} | \vec{x}_{<}, \vec{x}_{t,i,j,R})\ p(\vec{x}_{t,i,j,R} | \vec{x}_{<}).
\end{equation}
Here $\vec{x}_{<}=\vec{x}_{(t,<i,<j,:)}\cup\vec{x}_{(<t,:,:,:)}$ comprises the RGB values of all pixels to the left and above the pixel at position $(i,j)$ in the current frame $t$, as well as the RGB values of pixels from all the previous frames. 
  
Note that the factorization itself does not impose a unique ordering on the set of variables. We choose an ordering according to two criteria. The first criterion is determined by the properties and uses of the data; frames in the video are predicted according to their temporal order. The second criterion favors orderings that can be computed efficiently; pixels are predicted starting from one corner of the frame  (the top left corner)  and ending in the opposite corner of the frame (the bottom right one) as this allows for the computation to be implemented efficiently \citep{van2016pixel}. The order for the prediction of the colors is chosen by convention as R, G and B.

The VPN models directly the four dimensions of video tensors.  We use $F_t$ to denote the $t$-th frame $\vec{x}_{t,:,:,:}$ in the video $\vec{x}$.

Figure~\ref{fig:dependencies} illustrates the fourfold dependency structure for the green color channel value of the pixel $x$ in frame $F_t$,
which depends on:
(i) all pixels in all the previous frames $F_{<t}$;
(ii) all three colors of the already generated pixels in $F_t$;
(iii) the already generated red color value of the pixel $x$.

We follow the PixelRNN approach \citep{van2016pixel} in modelling each conditional factor as a discrete multinomial distribution over 256 raw color values. 
This allows for the predicted distributions to be arbitrarily multimodal.

\subsection{Baseline Model}

We compare the VPN model with a baseline model that encodes the temporal dependencies in videos from previous frames to the next, but ignores the spatial dependencies between the pixels within a frame and the dependencies between the color channels. In this case the joint distribution is factorized by introducing independence assumptions:
\begin{equation}
p(\vec{x}) \approx \prod_{t=0}^{T}\prod_{i=0}^N\prod_{j=0}^N p(\vec{x}_{t,i,j,B} | \vec{x}_{<t,:,:,:})\ p(\vec{x}_{t,i,j,G} | \vec{x}_{<t,:,:,:})\ p(\vec{x}_{t,i,j,R} | \vec{x}_{<t,:,:,:}).
\end{equation}
Figure \ref{fig:dependencies} illustrates the conditioning structure in the baseline model. The green channel value of pixel $\mathbf{x}$ only depends on the values of pixels in previous frames. Various models have been proposed that are similar to our baseline model in that they capture the temporal dependencies only \citep{DBLP:journals/corr/RanzatoSBMCC14,ICML-2015-SrivastavaMS,DBLP:conf/nips/OhGLLS15}

\subsection{Remarks on the Factorization}
\label{sec:discussion}

To illustrate the properties of the two factorizations, suppose that a model needs to predict the value of a pixel $x$ and the value of the adjacent pixel $y$ in a frame $F$, where the transition to the frame $F$ from the previous frames $F_<$ is non-deterministic. For a simple example, suppose the previous frames $F_<$ depict a robotic arm and in the current frame $F$ the robotic arm is about to move either left or right. The baseline model estimates $p(x|F_<)$ and $p(y|F_<)$ as distributions with two modes, one for the robot moving left and one for the robot moving right. Sampling independently from $p(x|F_<)$ and $p(y|F_<)$ can lead to two inconsistent pixel values coming from distinct modes, one pixel value depicting the robot moving left and the other depicting the robot moving right. The accumulation of these inconsistencies for a few frames leads to known artifacts such as blurring of video continuations. By contrast, in this example, the VPN estimates $p(x|F_<)$ as the same bimodal distribution, but then estimates $p(y|x, F_<)$ \emph{conditioned on the selected value of} $x$. The conditioned distribution is unimodal and, if the value of $x$ is sampled to depict the robot moving left, then the value of $y$ is sampled accordingly to also depict the robot moving left. 

Generating a video tensor requires sampling $T*N^2*3$ variables, which for a second of video with resolution $64 \times 64$ is in the order of $10^5$ samples. This figure is in the order of $10^4$ for generating a single image or for a second of audio signal \citep{wavenet}, and it is in the order of $10^2$ for language tasks such as machine translation \citep{DBLP:conf/emnlp/KalchbrennerB13}.

\section{Architecture}
\label{sec:arch}

In this section we construct a network architecture capable of computing efficiently the factorized distribution in~\sref{sec:factor}. The architecture consists of two parts. The first part models the temporal dimension of the data and consists of Resolution Preserving CNN Encoders whose outputs are given to a Convolutional LSTM. The second part models the spatial and color dimensions of the video and consists of PixelCNN architectures \citep{van2016pixel,DBLP:journals/corr/OordKVEGK16} that are conditioned on the outputs of the CNN Encoders.

\subsection{Resolution Preserving CNN Encoders}
Given a set of video frames $F_{0},...,F_{T}$, the VPN first encodes each of the first $T$ frames $F_0,...,F_{T-1}$ with a CNN Encoder. These frames form the histories that condition the generated frames. Each of the CNN Encoders is composed of $k$ ($k=8$ in the experiments) residual blocks (Sect.~\ref{sec:impl_details}) and the spatial resolution of the input frames is preserved throughout the layers in all the blocks. Preserving the resolution is crucial as it allows the model to condition each pixel that needs to be generated without loss of representational capacity. The outputs of the~$T$ CNN Encoders, which are computed in parallel during training, are given as input to a Convolutional LSTM, which also preserves the resolution. This part of the VPN computes the temporal dependencies of the video tensor and is represented in \figref{fig:dependencies} by the shaded blocks.

\subsection{PixelCNN Decoders}
\label{sec:temp_stream}
The second part of the VPN architecture computes dependencies along the space and color dimensions. The $T$ outputs of the first part of the architecture provide representations for the contexts that condition the generation of a portion of the $T+1$ frames $F_0,...,F_{T}$; if one generates all the $T+1$ frames, then the first frame $F_0$ receives no context representation. These context representations are used to condition decoder neural networks that are PixelCNNs. PixelCNNs are composed of $l$ resolution preserving residual blocks ($l=12$ in the experiments), each in turn formed of \emph{masked} convolutional layers. Since we treat the pixel values as discrete random variables, the final layer of the PixelCNN decoders is a softmax layer over 256 intensity values for each color channel in each pixel. 

Figure~\ref{fig:dependencies} depicts the two parts of the architecture of the VPN. The decoder that generates pixel $x$ of frame $F_t$ sees the context representation for all the frames up to $F_{t-1}$ coming from the preceding CNN encoders. The decoder also sees the pixel values above and left of the pixel $x$ in the current frame $F_t$ that is itself given as input to the decoder.

\subsection{Architecture of Baseline Model}

We implement the baseline model by using the same CNN encoders to build the context representations. In contrast with PixelCNNs, the decoders in the baseline model are CNNs that do not use masking on the weights; the frame to be predicted thus cannot be given as input. As shown in \figref{fig:dependencies}, the resulting neural network captures the temporal dependencies, but ignores spatial and color channel dependencies within the generated frames. Just like for VPNs, we make the neural architecture of the baseline model resolution preserving in all the layers.

\begin{figure}
\vspace{-1cm}
\centering
\includegraphics[width=0.5\linewidth]{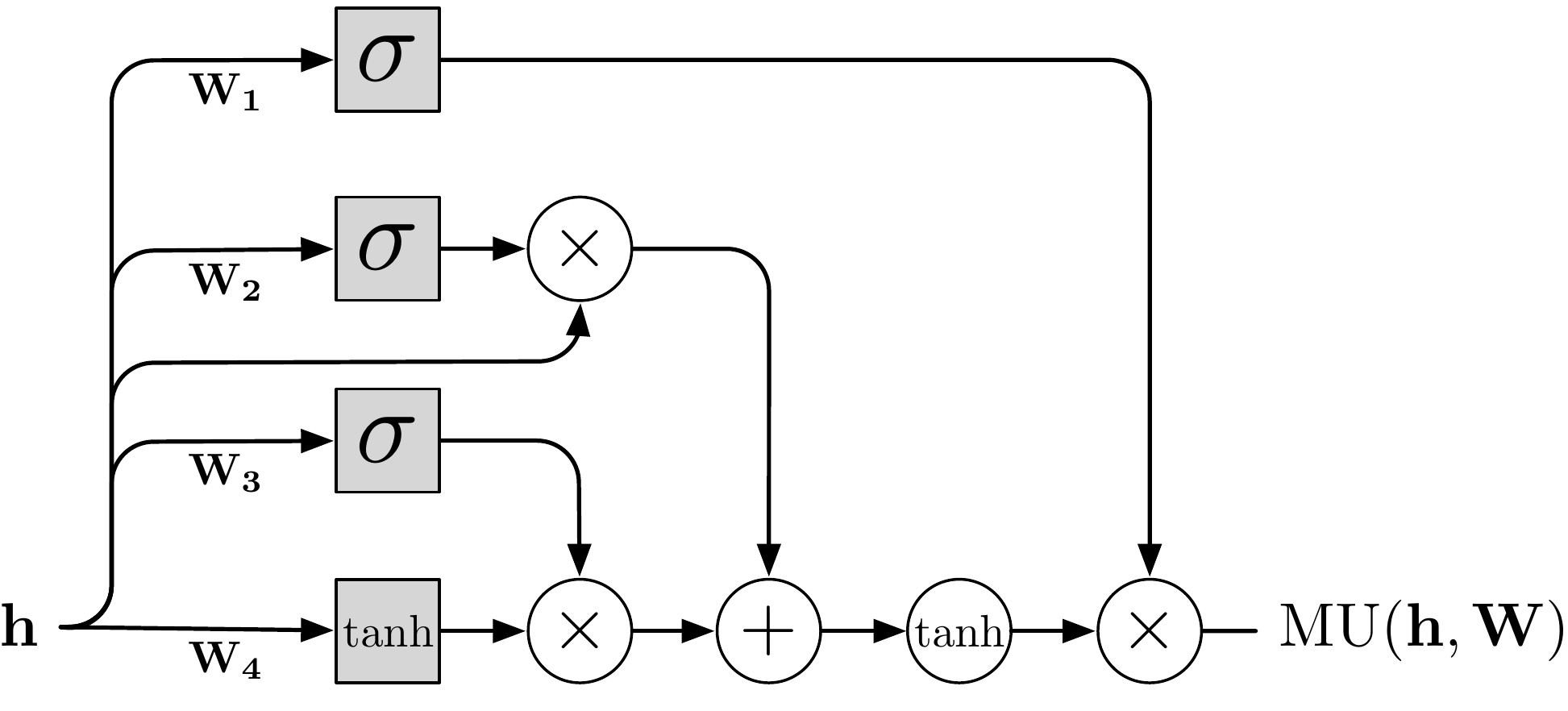}
\caption{Structure of a multiplicative unit (MU). The squares represent the three gates and the update. The circles represent component-wise operations.}
\label{fig:mu}
\end{figure}

\section{Network Building Blocks}
\label{sec:impl_details}

In this section we describe two basic operations that are used as the building blocks of the VPN. 
The first is the \emph{Multiplicative Unit} (MU,~\sref{sec:mu}) that contains multiplicative interactions 
inspired by LSTM~\citep{hochreiter1997long} gates. 
The second building block is the \emph{Residual Multiplicative Block} (RMB,~\sref{sec:rmb}) that is composed of multiple layers of MUs. 

\subsection{Multiplicative Units}
\label{sec:mu}

A multiplicative unit (\figref{fig:mu}) is constructed by incorporating LSTM-like gates into a convolutional layer.
Given an input $\vec{h}$ of size $N \times N \times c$, where $c$ corresponds to the number of channels, we first pass it through four convolutional layers to create an update $\vec{u}$ and three gates $\vec{g_{1-3}}$. The input, update, and gates are then combined in the following manner:

\begin{align}
\nonumber&\vec{g_1} = \sigma (\mathbf{W_1 \ast h}) \\
\nonumber&\vec{g_2} = \sigma (\mathbf{W_2 \ast h}) \\
&\vec{g_3} = \sigma (\mathbf{W_3 \ast h}) \\
\nonumber&\vec{u} = \tanh (\mathbf{W_4 \ast h}) \\
\nonumber\mbox{MU}(\mathbf{h};\vec{W}) &= \vec{g_1} \odot \tanh( \vec{g_2} \odot \mathbf{h} + \vec{g_3} \odot \vec{u} ) 
\end{align}

where $\sigma$ is the sigmoid non-linearity and $\odot$ is component-wise multiplication. Biases are omitted for clarity.
In our experiments the convolutional weights $\vec{W_{1-4}}$ use a kernel of size $3\times3$.
Unlike LSTM networks, there is no distinction between \emph{memory} and \emph{hidden} states. 
Also, unlike Highway networks \citep{DBLP:journals/corr/SrivastavaGS15} and Grid LSTM \citep{DBLP:journals/corr/KalchbrennerDG15}, there is no setting of the gates such that $\mbox{MU}(\mathbf{h};\vec{W})$ simply returns the input $\vec{h}$; the input is always processed with a non-linearity.

\begin{figure}[t]
\vspace{-0.5cm}
\centering
\includegraphics[width=0.9\linewidth]{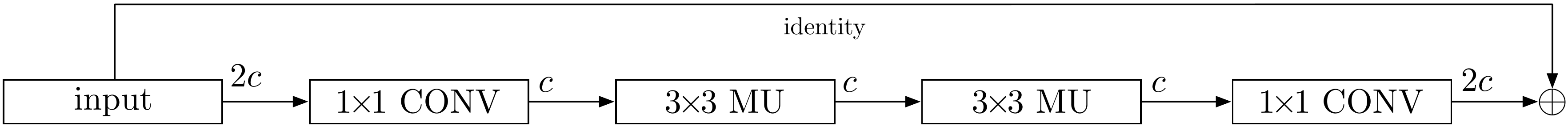}
\caption{Structure of the residual multiplicative block (RMB) incorporating two multiplicative units (MUs).}
\label{fig:rmb}
\end{figure}

\subsection{Residual Multiplicative Blocks}
\label{sec:rmb}
To allow for easy gradient propagation through many layers of the network, we stack two MU layers in a residual multiplicative block (\figref{fig:rmb}) where the input has a residual (additive skip) connection to the output \citep{DBLP:journals/corr/HeZR016}. For computational efficiency, the number of channels is halved in MU layers inside the block.
Namely, given an input layer $\vec{h}$ of size $N \times N \times 2c$ with $2c$ channels, we first apply a $1 \times 1$ convolutional layer that reduces the number of channels to $c$;
no activation function is used for this layer, and it is followed by two successive MU layers each with a convolutional kernel of size $3\times3$. 
We then project the feature map back to $2c$ channels using another $1\times 1$ convolutional layer. Finally, the input $\vec{h}$ is added to the overall output forming a residual connection.
Such a layer structure is similar to the bottleneck residual unit of~\citep{DBLP:journals/corr/HeZR016}.
Formally, the Residual Multiplicative Block (RMB) is computed as follows:
\begin{align}
\nonumber&\vec{h_1} =  \mathbf{W_1 \ast h} \\
\nonumber&\vec{h_2} = \mbox{MU} (\vec{h_1};\vec{W_2}) \\
&\vec{h_3} = \mbox{MU} (\vec{h_2};\vec{W_3}) \\
\nonumber&\vec{h_4} =  \mathbf{W_4 \ast h_3} \\
\nonumber&\mbox{RMB}(\mathbf{h};\vec{W}) = \mathbf{h} + \vec{h_4} 
\end{align}

We also experimented with a standard residual block of~\citep{DBLP:journals/corr/HeZR016} which uses ReLU non-linearities -- see~\sref{sec:moving_mnist} and~\ref{sec:robotic_pushing} for details.

\subsection{Dilated Convolutions}
Having a large receptive field helps the model to capture the motion of larger objects. One way to increase the receptive field without much effect on the computational complexity is to use dilated convolutions~\citep{DBLP:journals/corr/ChenPKMY14,DBLP:journals/corr/YuK15}, which make the receptive field grow exponentially, as opposed to linearly, in the number of layers. In the variant of VPN that uses dilation, the dilation rates are the same within each RMB, but they double from one RMB to the next up to a chosen maximum size, and then repeat~\citep{wavenet}. In particular, in the CNN encoders we use two repetitions of the dilation scheme $\lbrack 1, 2, 4, 8 \rbrack$, for a total of 8 RMBs. We do not use dilation in the decoders.

\section{Moving MNIST}
\label{sec:moving_mnist}

\begin{table}[t]
\small
  \begin{center}
  \begin{tabular}{lccc}
    \toprule
    \textbf{Model} &  \textbf{Test}  \\ \midrule
      \citep{DBLP:conf/nips/ShiCWYWW15} & 367.2 \\
      \citep{ICML-2015-SrivastavaMS} & 341.2 \\ 
      \citep{DBLP:journals/corr/BrabandereJTG16} & 285.2 \\ 
      \citep{DBLP:journals/corr/PatrauceanHC15} & 179.8 \\ 
      {Baseline model} & 110.1  \\ 
      \textbf{VPN} &  $\mathbf{87.6}$ \\ \midrule
      {Lower Bound} & 86.3  \\
      \bottomrule
  \end{tabular}
  \end{center}
\caption{Cross-entropy results in nats/frame on the Moving MNIST dataset. }
\label{table:mnist}
\end{table}

%
The Moving MNIST dataset consists of sequences of 20 frames of size $64 \times 64$, depicting two potentially overlapping MNIST digits moving with constant velocity and bouncing off walls. Training sequences are generated on-the-fly using digits from the MNIST training set without a limit on the number of generated training sequences (our models observe 19.2M training sequences before convergence). The test set is fixed and consists of 10000 sequences that contain digits from the MNIST test set. 10 of the 20 frames are used as context and the remaining 10 frames are generated. 

In order to make our results comparable, for this dataset only we use the same sigmoid cross-entropy loss as used in prior work \citep{ICML-2015-SrivastavaMS}. The loss is defined as:
\begin{equation}
H(z,y) = -\sum_{i}z_i \log y_i + (1-z_i)\log(1-y_i)
\end{equation}
where $z_i$ are the grayscale targets in the Moving MNIST frames that are interpreted as probabilities and $y_i$ are the predictions. The lower bound on $H(z,y)$ may be non-zero. In fact, if we let $z_i=y_i$, for the 10 frames that are predicted in each sequence of the Moving MNIST test data, $H(z,y) = 86.3$ nats/frame.

\subsection{Implementation Details}

The VPNs with and without dilation, as well as the baseline model, have 8 RMBs in the encoders and 12 RMBs in the decoders; for the network variants that use ReLUs we double the number of residual blocks to 16 and 24, respectively, in order to equate the size of the receptive fields in the two model variants. The number of channels in the blocks is $c=128$ while the convolutional LSTM has 256 channels. The topmost layer before the output has 768 channels. We train the models for 300000 steps with 20-frame sequences predicting the last 10 frames of each sequence.  Each step corresponds to a batch of 64 sequences. We use RMSProp for the optimization with an initial learning rate of $3 \cdot 10^{-4}$ and multiply the learning rate by $0.3$ when learning flatlines.  

\begin{table}[t]
\small
  \begin{center}
  \begin{tabular}{lccc}
    \toprule
    \textbf{Model} &  \textbf{Test}  \\ \midrule
      {VPN (RMB, No Dilation)} &  89.2 \\ 
      {VPN (Relu, No Dilation)} &  89.1 \\ 
      {VPN (Relu, Dilation)} &  87.7 \\ 
      {{VPN (RMB, Dilation)}} &  87.6 \\ \midrule
      {Lower Bound} & 86.3  \\
      \bottomrule
  \end{tabular}
  \end{center}
\caption{Cross-entropy results in nats/frame on the Moving MNIST dataset.}
\label{table:dilation}
\end{table}

\subsection{Results}

Table~\ref{table:mnist} reports the results of various recent video models on the Moving MNIST test set. Our baseline model achieves 110.1 nats/frame, which is  significantly better than the previous state of the art~\citep{DBLP:journals/corr/PatrauceanHC15}. We attribute these gains to architectural features and, in particular, to the resolution preserving aspect of the network. Further, the VPN achieves 87.6 nats/frame, which approaches the lower bound of 86.3 nats/frame. 

Table~\ref{table:dilation} reports results of architectural variants of the VPNs. The model with dilated convolutions improves over its non-dilated counterpart as it can more easily act on the relatively large digits moving in the $64 \times 64$ frames. In the case of Moving MNIST, MUs do not yield a significant improvement in performance over just using ReLUs, possibly due to the relatively low complexity of the task. A sizeable improvement is obtained from MUs on the Robotic Pushing dataset (Tab.~\ref{table:robot_act}).

\begin{figure}
\vspace{-2cm}
\hspace{-1.5cm}
\includegraphics[width=1.2\linewidth]{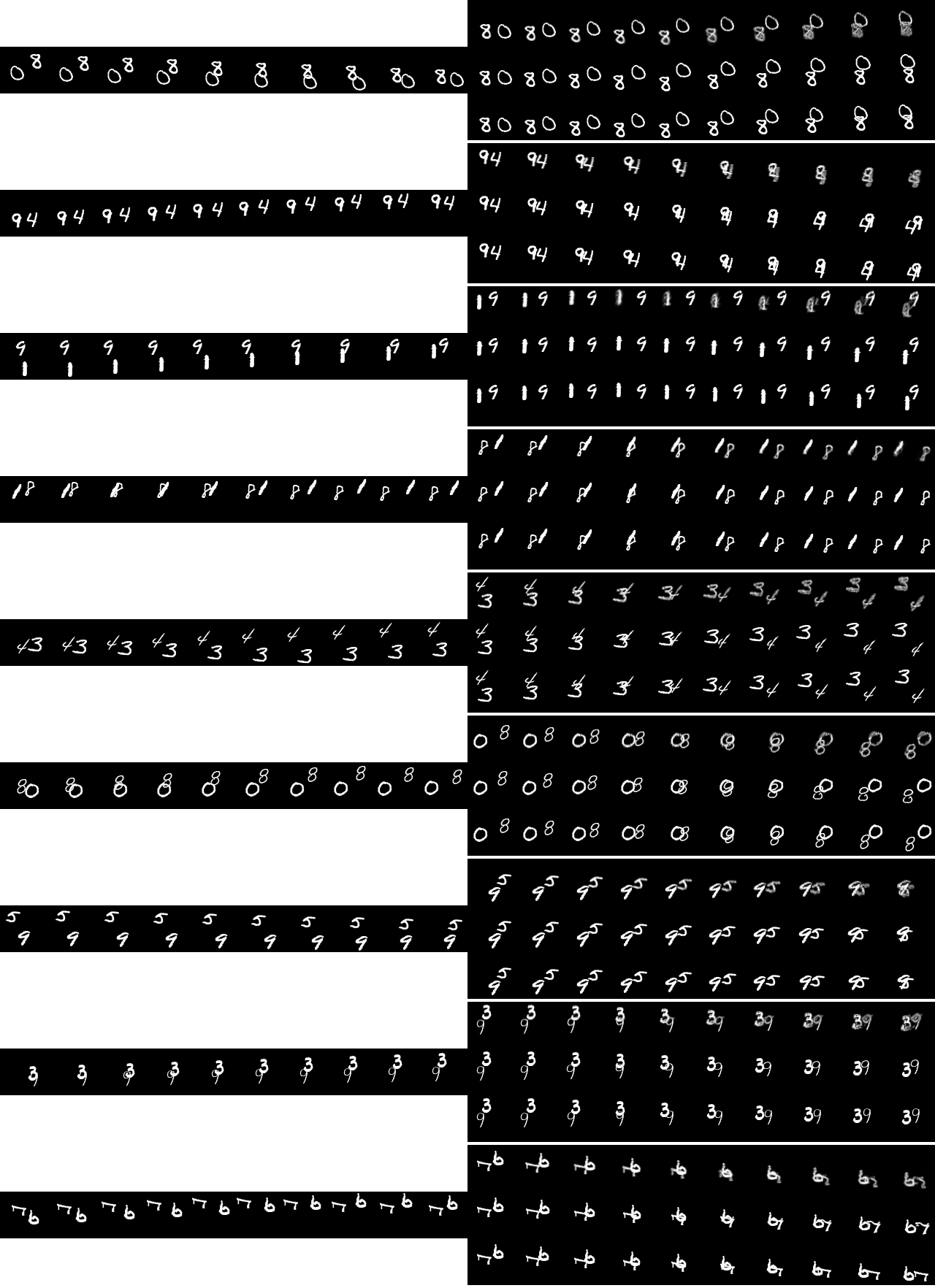}
\caption{Randomly sampled continuations of videos from the Moving MNIST test set. For each set of three rows, the first 10 frames in the middle row are the given context frames. The next three rows of 10 frames each are as follows: frames generated from the baseline model (top row), frames generated from the VPN (middle row) and ground truth frames (bottom row).}
\label{fig:samples_mnist}
\end{figure}

A qualitative evaluation of video continuations produced by the models matches the quantitative results. Figure~\ref{fig:samples_mnist} shows random continuations produced by the VPN and the baseline model on the Moving MNIST test set. The frames generated by the VPN are consistently sharp even when they deviate from the ground truth.

By contrast, the continuations produced by the baseline model get progressively more blurred with time -- as the uncertainty of the model grows with the number of generated frames, the lack of inter-frame spatial dependencies leads the model to take the expectation over possible trajectories.

\section{Robotic Pushing}
\label{sec:robotic_pushing}

The Robotic Pushing dataset consists of sequences of 20 frames of size $64 \times 64$ that represent camera recordings of a robotic arm pushing objects in a basket. The data consists of a training set of 50000 sequences, a validation set, and two test sets of 1500 sequences each, one involving a subset of the objects seen during training and the other one involving \emph{novel} objects not seen during training. Each frame in the video sequence is paired with the state of the robot at that frame and with the desired action to reach the next frame. The transitions are non-deterministic as the robotic arm may not reach the desired state in a frame due to occlusion by the objects encountered on its trajectory. 2 frames, 2 states and 2 actions are used as context; the desired 18 actions in the future are also given. The 18 frames in the future are then generated conditioned on the 18 actions as well as on the 2 steps of context. 

\subsection{Implementation Details}

For this dataset, both the VPN and the baseline model use the softmax cross-entropy loss, as defined in Sect.~\ref{sec:factor}. As for Moving MNIST, the models have 8 RMBs in the encoders and 12 RMBs in the decoders; the ReLU variants have 16 residual blocks in the encoders and 24 in the decoders. The number of channels in the RMBs is $c=128$, the convolutional LSTM has 256 channels and the topmost layer before the output has 1536 channels. We use RMSProp with an initial learning rate of $10^{-4}$. We train for 275000 steps with a batch size of 64 sequences per step. Each training sequence is obtained by selecting a random subsequence of 12 frames together with the corresponding states and actions. We use the first 2 frames in the subsequence as context and predict the other 10 frames. States and actions come as vectors of 5 real values. For the 2 context frames, we condition each layer in the encoders and the decoders with the respective state and action vectors; the conditioning is performed by the result of a $1 \times 1$ convolution applied to the action and state vectors that are broadcast to all of the $64 \times 64$ positions.  For the other 10 frames, we condition the encoders and decoders with the action vectors only. We discard the state vectors for the predicted frames during training and do not use them at generation. For generation we unroll the models for the entire sequence of 20 frames and generate 18 frames.

\begin{table}[t]
\small
  \begin{center}
  \small
  \begin{tabular}{lccc}
    \toprule
    \textbf{Model} &  \textbf{Validation} &  \textbf{Test (Seen)} &  \textbf{Test (Novel)} \\ \midrule
      {Baseline model} & 2.06 & 2.08 & 2.07 \\ 
      {VPN (Relu, Dilation)} &  0.73 & 0.72 & 0.75 \\ 
      {VPN (Relu, No Dilation)} &  0.72 & 0.73 & 0.75 \\ 
      {VPN (RMB, Dilation)} &  0.63 & 0.65 & 0.64 \\ 
      {\textbf{VPN} (RMB, No Dilation)} &  $\mathbf{0.62}$ & $\mathbf{0.64}$ & $\mathbf{0.64}$  \\
      \bottomrule
  \end{tabular}
  \end{center}
\caption{Negative log-likelihood in nats/dimension on the Robotic Pushing dataset. }
\label{table:robot_act}
\end{table}

\subsection{Results}

Table~\ref{table:robot_act} reports the results of the baseline model and variants of the VPN on the Robotic Pushing validation and test sets. The best variant of the VPN has a $>65\%$ reduction in negative log-likelihood over the baseline model. This highlights the importance of space and color dependencies in non-deterministic environments. The results on the validation and test datasets with seen objects and on the test dataset with novel objects are similar. This shows that the models have learned to generalize well not just to new action sequences, but also to new objects. Furthermore, we see that using multiplicative interactions in the VPN gives a significant improvement over using ReLUs.

\begin{figure}
\vspace{-2cm}
\includegraphics[width=\linewidth]{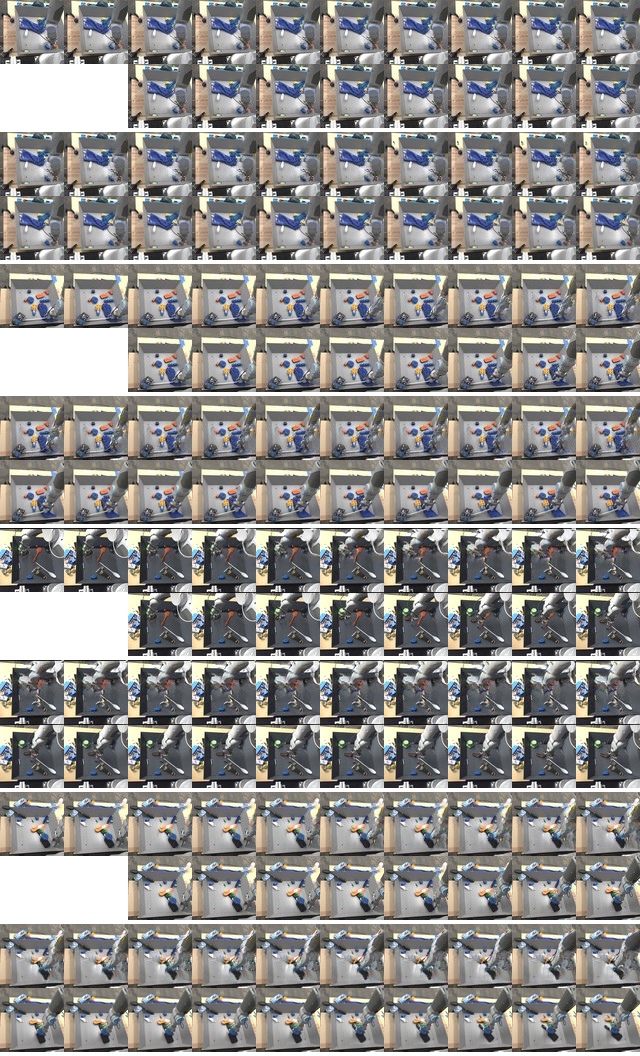}
\caption{Randomly sampled continuations of videos from the Robotic Pushing validation set (with seen objects). Each set of four rows corresponds to a sample of 2 given context frames and 18 generated frames. In each set of four rows, rows 1 and 3 are samples from the VPN. Rows 2 and 4 are the actual continuation in the data.}
\label{fig:robot_valid}
\end{figure}

\begin{figure}

\includegraphics[width=\linewidth]{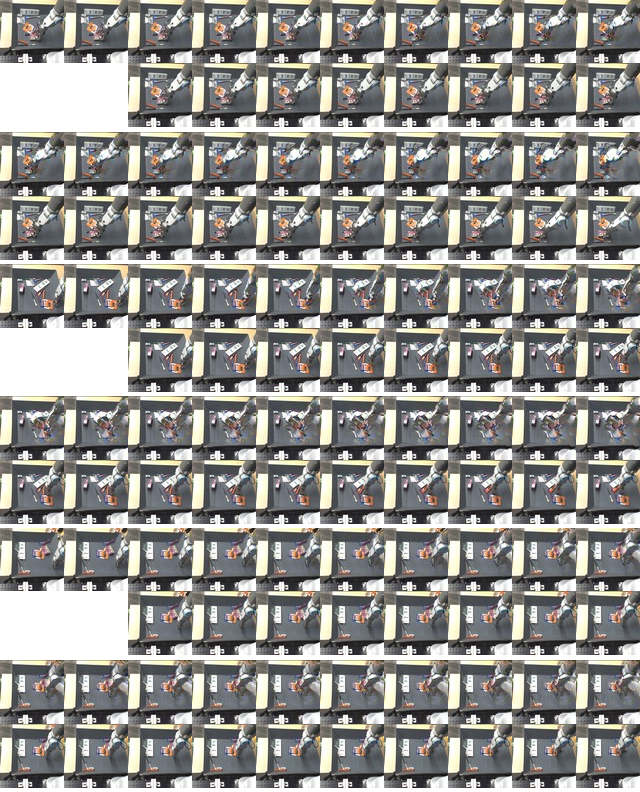}
\caption{Randomly sampled continuations of videos from the Robotic Pushing test set with \emph{novel} objects not seen during training. Each set of four rows is as in \figref{fig:robot_valid}. }
\label{fig:robot_novel}
\end{figure}

\begin{figure}

\includegraphics[width=\linewidth]{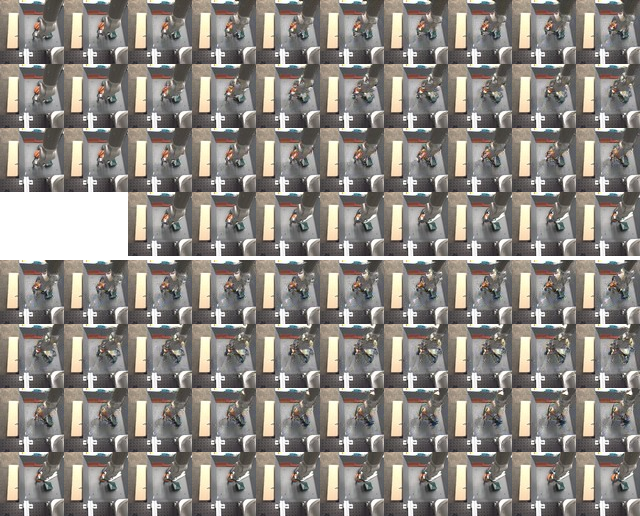}
\caption{Three different samples from the VPN starting from the same 2 context frames on the Robotic Pushing validation set. For each set of four rows, top three rows are generated samples, the bottom row is the actual continuation in the data.}
\label{fig:robot_cont}
\end{figure}

\begin{figure}
\includegraphics[width=1\linewidth]{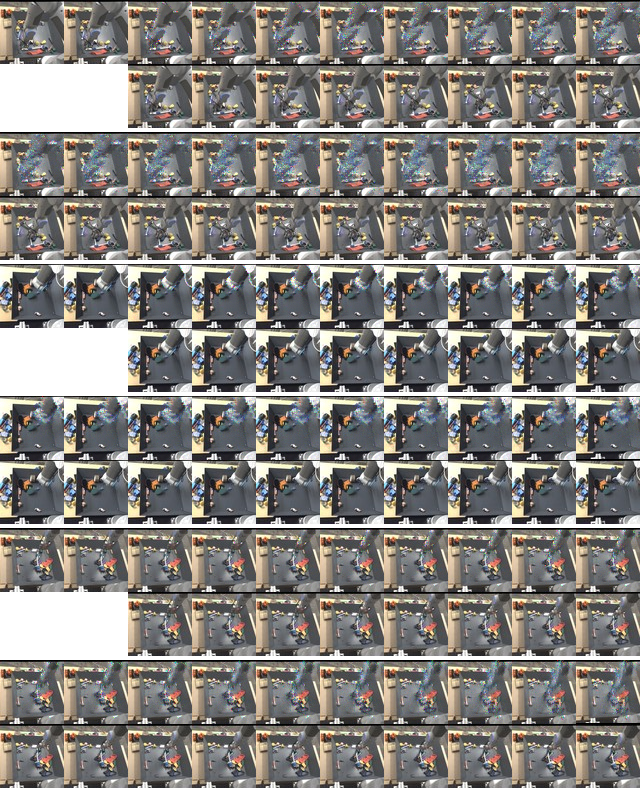}
\caption{Randomly sampled continuations from the baseline model on the Robotic Pushing validation set (with seen objects). Each set of four rows is as in \figref{fig:robot_valid}. }
\label{fig:robot_base}
\end{figure}

\begin{figure}

\includegraphics[width=1\linewidth]{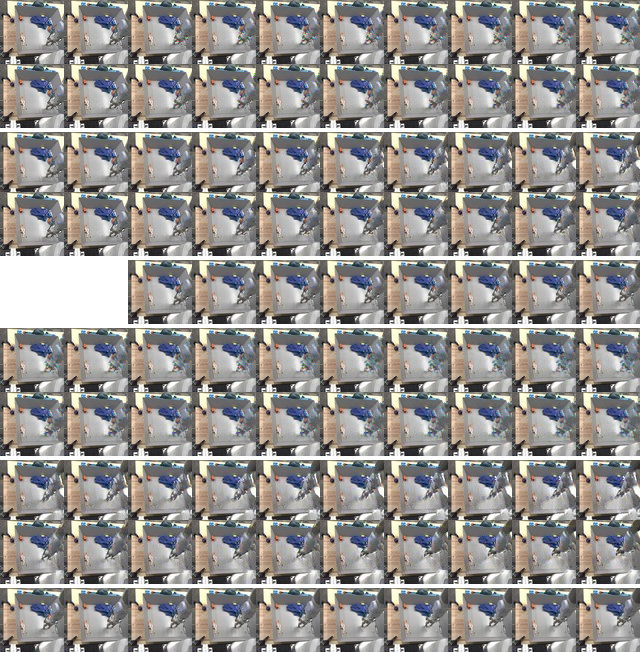}
\caption{Comparison of continuations given the same 2 context frames from the Robotic Pushing validation set for the baseline model (rows 1 and 2, and 6 and 7), for the VPN (rows 3 and 4, and 8 and 9) and for the actual continuation in the data (rows 5 and 10).}
\label{fig:robot_comp}
\end{figure}

Figures \ref{fig:robot_valid}--\ref{fig:robot_comp} visualize the samples generated by our models. Figure \ref{fig:robot_valid} contains random samples of the VPN on the validation set with seen objects (together with the corresponding ground truth). The model is able to distinguish between the robotic arm and the background, correctly handling occlusions and only pushing the objects when they come in contact with the robotic arm. The VPN generates the arm when it enters into the frame from one of the sides. The position of the arm in the samples is close to that in the ground truth, suggesting the VPN has learned to follow the actions. The generated videos remain detailed throughout the 18 frames and few artifacts are present.  The samples remain good showing the ability of the VPN to generalize to new sequences of actions.  Figure \ref{fig:robot_novel} evaluates an additional level of generalization, by showing samples from the test set with \emph{novel} objects not seen during training. The VPN seems to identify the novel objects correctly and generates plausible movements for them. The samples do not appear visibly worse than in the datasets with seen objects. Figure \ref{fig:robot_cont} demonstrates the probabilistic nature of the VPN, by showing multiple different video continuations that start from the same context frames and are conditioned on the same sequence of 18 future actions. The continuations are plausible and varied, further suggesting the VPN's ability to generalize. Figure \ref{fig:robot_base} shows samples from the baseline model. In contrast with the VPN samples, we see a form of high frequency noise appearing in the non-deterministic movements of the robotic arm. This can be attributed to the lack of space and color dependencies, as discussed in Sec.~\ref{sec:discussion}. Figure \ref{fig:robot_comp} shows a comparison of continuations of the baseline model and the VPN from the same context sequence. Besides artifacts, the baseline model also seems less responsive to the actions.

\section{Conclusion}

We have introduced the Video Pixel Network, a deep generative model of video data that models the factorization of the joint likelihood of video. We have shown that, despite its lack of specific motion priors or surrogate losses, the VPN approaches the lower bound on the loss on the Moving MNIST benchmark that corresponds to a large improvement over the previous state of the art. On the Robotic Pushing dataset, the VPN achieves significantly better likelihoods than the baseline model that lacks the fourfold dependency structure; the VPN generates videos that are free of artifacts and are highly detailed for many frames into the future. The fourfold dependency structure provides a robust and generic method for generating videos without systematic artifacts.

\subsubsection*{Acknowledgments}

The authors would like to thank Chelsea Finn for her advice on the Robotic Pushing dataset and Lasse Espeholt for helpful remarks.

\bibliographystyle{plainnat}
{
\small
\bibliography{main}
}

\centering

\end{document}